%% file: main.tex
\pdfoutput=1
\documentclass[letterpaper]{article}
\usepackage[preprint]{aaai2027}
\usepackage[hyphens]{url}
\usepackage{graphicx}
\usepackage{natbib}
\usepackage{caption}
\usepackage{amssymb}
\usepackage{booktabs}
\usepackage{algorithm}
\usepackage{algorithmic}
\usepackage{tikz}
\usetikzlibrary{arrows.meta}
\usepackage{colortbl}
\usepackage[most]{tcolorbox}
\tcbuselibrary{listings}
\definecolor{promptframe}{RGB}{47,68,102}
\definecolor{promptbg}{RGB}{245,247,250}
\definecolor{jsonstring}{RGB}{146,104,10}
\definecolor{jsonkw}{RGB}{27,94,63}
\newtcolorbox{promptbox}[2][]{%
  enhanced, breakable,
  colback=promptbg, colframe=promptframe,
  coltitle=white, fonttitle=\bfseries\small,
  title={#2},
  boxrule=0.6pt, arc=1.5pt,
  left=5pt, right=5pt, top=4pt, bottom=4pt,
  before upper={\small\ttfamily\raggedright}, #1}
\lstdefinelanguage{json}{
  morestring=[b]",
  morekeywords={true,false,null},
  sensitive=true}
\lstdefinestyle{jsonstyle}{
  language=json,
  basicstyle=\scriptsize\ttfamily\color{promptframe},
  stringstyle=\color{jsonstring},
  keywordstyle=\color{jsonkw},
  breaklines=true,
  breakatwhitespace=false,
  postbreak=\mbox{\hspace{1em}},
  showstringspaces=false,
  keepspaces=true,
  columns=flexible}
\newtcblisting{jsonbox}[1]{%
  enhanced, breakable=false,
  colback=promptbg, colframe=promptframe,
  coltitle=white, fonttitle=\bfseries\small,
  title={#1},
  boxrule=0.6pt, arc=1.5pt,
  left=5pt, right=5pt, top=4pt, bottom=4pt,
  listing only,
  listing options={style=jsonstyle}}
\newenvironment{supptable}{%
  \par\smallskip\noindent\begin{minipage}{\columnwidth}%
  \captionsetup{type=table}\centering\small}{%
  \end{minipage}\par\smallskip}
\definecolor{ocdiff}{RGB}{233,237,242}
\definecolor{ocmain}{RGB}{32,105,68}
\definecolor{ocwarn}{RGB}{176,57,46}
\definecolor{ocflag}{RGB}{158,108,20}
\definecolor{ocink}{RGB}{39,46,52}
\definecolor{ocmuted}{RGB}{105,111,116}
\definecolor{ocgrid}{RGB}{222,226,229}
\definecolor{ocbluewash}{RGB}{240,247,242}
\definecolor{ocwarmwash}{RGB}{252,247,246}
\definecolor{ocamberwash}{RGB}{253,247,229}
\title{Outcome Monitors: Recovery Affordances for Silent Tool Failures}
\author{
    Sugam Panthi,
    Rabab Abdelfattah
}
\affiliations{
    The University of Southern Mississippi\\
    sugam.panthi@usm.edu, rabab.abdelfattah@usm.edu
}

\newcommand{\method}{Outcome Monitors}
\newcommand{\pp}{percentage points}
\newcommand{\icondeepseek}{\raisebox{-0.15ex}{\includegraphics[height=1.1em]{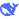}}}
\newcommand{\iconqwen}{\raisebox{-0.15ex}{\includegraphics[height=1.1em]{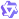}}}
\newcommand{\iconminimax}{\raisebox{-0.15ex}{\includegraphics[height=1.1em]{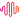}}}

\begin{document}

\maketitle

\input{sections/01-abstract}
\input{sections/02-introduction}

\input{sections/03-related-work}
\input{sections/04-methodology}
\input{sections/05-experimental-design}
\input{sections/06-results}
\input{sections/07-discussion}
\input{sections/08-conclusion}

\bibliography{references}

\clearpage
\appendix
\begin{center}
{\LARGE\bfseries Supplementary Material: \method}\par
\vspace{0.5em}
{\large Outcome Monitors: Recovery Affordances for Silent Tool Failures}\par
\end{center}
\vspace{1em}
\input{sections/supplement}

\end{document}

%% file: sections/01-abstract.tex
\begin{abstract}
When a tool call times out, the agent sees the failure and can route around it.
A cached error page or negative price can instead arrive in the expected
format and be consumed as fact. We introduce\textbf{ \method}, which detect
violations of outcome contracts mined from task-disjoint traces or derived
from public schemas. On a violation, the monitor preserves the result and issues a
nonbinding receipt naming the violated property and public recovery tools.
In frozen, prespecified evaluations with injected failures, \method{} raise
ToolMaze completion from $10.9\%$ to $28.1\%$ across four models in two
provider families and replicate in a third. In $\tau$-bench retail, completion
improves by $14.0$ and $12.0$ points on two tiers. In separate ToolMaze
controls, removing the recovery-tool list eliminates the measured gain and
restoring it recovers the effect; diagnostic detail and timing produce no
detectable differences. Gains concentrate where the fault blocks completion.
On a suite transcribed from a published incident taxonomy, detection outside
the mined vocabulary falls to $46\%$, though delivery continues and completion
is unchanged. Recovery tools are the active receipt content in these controls;
extending detection beyond the contract vocabulary remains open.
\end{abstract}

%% file: sections/02-introduction.tex
\section{Introduction}

Language agents frequently combine reasoning with external tool calls~\cite{yao2023react,schick2023toolformer}. This lets them perform more tasks, but it also makes them rely on information they cannot directly observe. Timeouts are easy to detect. Contradictory, stale, or physically impossible results might instead arrive in the expected format and be taken at face value: in one documented production incident, a cached HTTP error page consumed as content became a confident fabricated analysis, and 70\% of that runtime's silent failures were first caught by a human reading output~\cite{wu2026errors}. ToolMaze demonstrates that agents struggle to recover from such implicit failures~\cite{zhu2026toolmaze}.

Silent failure creates two distinct reasoning problems. The system must first recognize that an observation is inconsistent and localize the violated property. Then, it must select a recovery action that advances the task. Traditional agent loops rely on the language model to infer both from the raw observation. At the opposite extreme, a runtime guard prevents actions or selects repairs, obscuring whether improved behavior stems from better reasoning or the guard's intervention.

We introduce a deterministic detector that identifies when a tool result violates a mined or public invariant and signals the violation to the acting agent. We name these invariants \emph{outcome contracts}, as they specify expected relationships between tool calls and their results, mined from nominal executions or derived from public schemas. On a violation the detector appends an advisory receipt; it never restricts an action, executes a repair, reveals a benchmark fault label, or consults the evaluator.

Recent contract and gate systems attach checks before execution and enforce them~\cite{liu2026toolgate,reddy2026reason,bhardwaj2026behavioral}. Our monitor instead checks a returned result and does not enforce a response. ToolMaze contracts use task-disjoint cross-fitting, and injection changes only what the model observes; evaluators therefore score the agent's chosen recovery rather than a repair encoded by the runtime.

The resulting interface separates detection from recovery: the monitor identifies a violated outcome property and lists public recovery tools, while the language agent chooses what to do. We call these tools the signal's \emph{recovery affordances}. Three findings support the interface:

\begin{enumerate}
\item \textbf{Detector-triggered receipts raise completion across
  three provider families.} We find paired completion gains with \method{} on a
  frozen, task-disjoint ToolMaze confirmation. The effect holds for four models
  spanning DeepSeek and Qwen. A separately frozen MiniMax M3 replication
  extends it to a third provider family.

\item \textbf{Benefits replicate in $\tau$-bench and concentrate where failures
  block completion.} We evaluate 50 $\tau$-bench retail tasks, each subject to both fault
  types. Categorical violations improve sharply with no paired losses, while
  conservation violations show no measured effect. Across all studies, the gains
  concentrate where the fault blocks baseline completion. Delivery is not
  universally beneficial: held-out AppWorld shows no measured net effect, and
  clean ToolMaze controls contain both rescues and harms.

\item \textbf{Recovery tools---not additional diagnostic detail---drive the
  detectable ToolMaze gain.}
  Neither the baseline's standing fault-aware prompt nor an extended
  reasoning budget removes the effect. A generic caution matches the
  localized witness, and changing placement or salience shows no
  detectable advantage. A warning stripped of recovery tools
  performs at baseline, while restoring the list restores the gain. A
  schema-only detector recovers much of the fault-enriched gain but
  fires more often; the learned contracts improve detection selectivity.
\end{enumerate}

We measure the remaining boundary with an incident-derived study. Outside the mined vocabulary, detection degrades but delivery does not.

\input{figures/fig1_detect_advise_tinted}

%% file: figures/fig1_detect_advise_tinted.tex
% Figure 1: raw interface vs detect-and-advise, single column, drawn at print size.
% Icons: Phosphor (light weight, MIT) -- see figures/icons/phosphor/PHOSPHOR-LICENSE.txt
% No \resizebox anywhere: every label below is its true printed point size.
% Sans labels deliberately separate the diagram's register from body Times.
% Row widths were measured with \settowidth, not estimated; the widest row
% (contract, 6.26cm) has 0.16cm of slack against the 7.82cm box interior.
\begin{figure}[t]
\centering
\begin{tikzpicture}[
  x=1cm, y=1cm,
  icon/.style={inner sep=0pt, outer sep=0pt},
  lane/.style={font=\sffamily\small\bfseries, anchor=north west, inner sep=0pt},
  txt/.style={font=\sffamily\footnotesize, text=ocink,
              anchor=west, inner sep=0pt},
  stepbox/.style={rounded corners=2pt, line width=0.65pt},
  flow/.style={draw=ocmuted, line width=1.15pt,
               -{Stealth[length=4.6pt, width=4.0pt]}},
]
\def\bl{0.00}\def\br{8.30}   % band left / right
\def\sl{0.32}\def\sr{7.98}   % step-box left / right
\def\ix{0.82}                % icon centre x  ==  arrow spine
\def\tx{1.40}                % text x

% ================================================================ lane A
\fill[ocwarmwash, rounded corners=3pt] (\bl,0.46) rectangle (\br,-3.50);
\node[lane, ocink] at (\sl,0.30) {RAW INTERFACE};

\draw[stepbox, fill=white, draw=ocgrid] (\sl,-0.56) rectangle (\sr,-1.20);
\node[icon] at (\ix,-0.88) {\includegraphics[height=14pt]{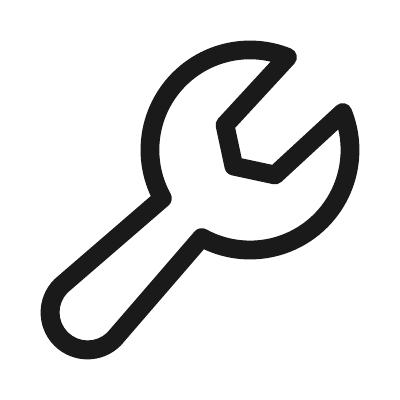}};
\node[txt] at (\tx,-0.88) {\ttfamily lookup\_inventory\quad\textcolor{ocmuted}{available = -3}};
\draw[flow] (\ix,-1.20) -- (\ix,-1.58);

\draw[stepbox, fill=white, draw=ocgrid] (\sl,-1.58) rectangle (\sr,-2.22);
\node[icon] at (\ix,-1.90) {\includegraphics[height=14pt]{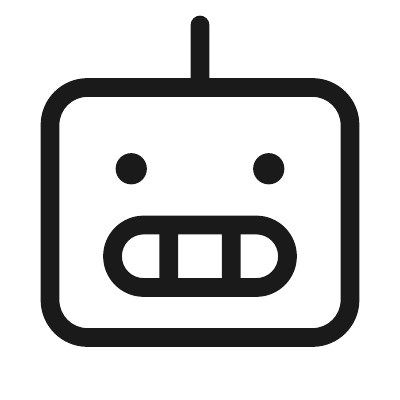}};
\node[txt] at (\tx,-1.90) {\textbf{detect \textperiodcentered{} localize \textperiodcentered{} recover}\quad\textcolor{ocmuted}{left to the agent}};
\draw[flow] (\ix,-2.22) -- (\ix,-2.60);

\draw[stepbox, fill=ocwarn!8, draw=ocwarn] (\sl,-2.60) rectangle (\sr,-3.24);
\node[icon] at (\ix,-2.92) {\includegraphics[height=14pt]{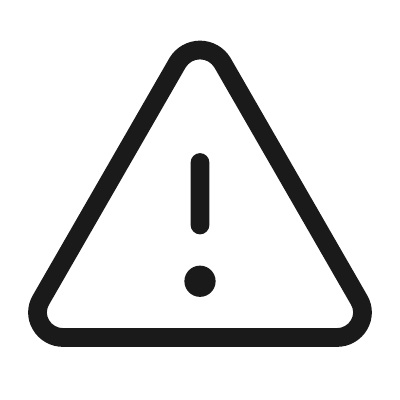}};
\node[txt] at (\tx,-2.92) {\textcolor{ocwarn}{\textbf{silent failure}}\quad\textcolor{ocmuted}{value consumed as fact}};

% ================================================================ lane B
\fill[ocbluewash, rounded corners=3pt] (\bl,-3.72) rectangle (\br,-9.90);
\node[lane, ocmain] at (\sl,-3.88) {DETECT AND ADVISE};

\draw[stepbox, fill=white, draw=ocgrid] (\sl,-4.58) rectangle (\sr,-5.22);
\node[icon] at (\ix,-4.90) {\includegraphics[height=14pt]{figures/icons/phosphor/wrench.pdf}};
\node[txt] at (\tx,-4.90) {\ttfamily lookup\_inventory\quad\textcolor{ocmuted}{raw result kept}};
\draw[flow] (\ix,-5.22) -- (\ix,-5.60);

\draw[stepbox, fill=ocamberwash, draw=ocflag] (\sl,-5.60) rectangle (\sr,-6.24);
\node[icon] at (\ix,-5.92) {\includegraphics[height=14pt]{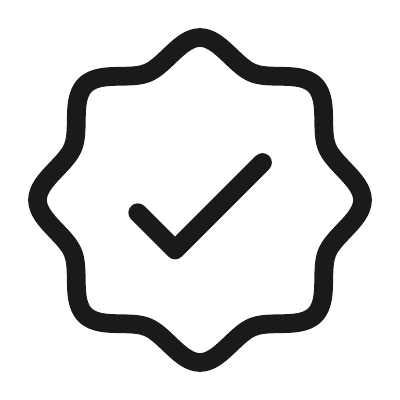}};
\node[txt] at (\tx,-5.92) {\textcolor{ocflag}{\textbf{contract violated}}\quad\textcolor{ocmuted}{mined or public invariant}};
\draw[flow] (\ix,-6.24) -- (\ix,-6.62);

% ---------------------------------------------------------------- receipt
\draw[fill=ocmain, draw=ocmain, line width=0.9pt, rounded corners=2.5pt]
  (\sl,-6.62) rectangle (\sr,-8.62);
\fill[white, rounded corners=1.5pt] (0.38,-7.18) rectangle (7.92,-8.56);
\node[font=\sffamily\footnotesize\bfseries, white, anchor=west] at (0.56,-6.90)
  {ADVISORY RECEIPT};
\node[font=\sffamily\footnotesize\bfseries, anchor=west] at (0.56,-7.48) {witness};
\node[font=\ttfamily\footnotesize, anchor=west] at (2.02,-7.48)
  {expected {$\geq$} 0, observed -3};
\node[font=\sffamily\footnotesize\bfseries, anchor=west] at (0.56,-7.92) {recovery};
\node[font=\ttfamily\footnotesize, anchor=west] at (2.02,-7.92)
  {query\_warehouse};
\node[font=\ttfamily\footnotesize, anchor=west] at (2.02,-8.28)
  {get\_stock\_by\_sku};
% the recovery list is the active content: brace it to the side, so the
% note cannot be misread as a third tool in the list
\draw[ocmain, line width=0.7pt] (5.36,-7.80) -- (5.36,-8.40);
\node[font=\sffamily\scriptsize\itshape, ocmain, anchor=west] at (5.52,-7.96)
  {active content};
\node[font=\sffamily\scriptsize\itshape, ocmain, anchor=west] at (5.52,-8.24)
  {(Section~\ref{sec:results})};
\draw[flow] (\ix,-8.62) -- (\ix,-9.00);

\draw[stepbox, fill=ocmain!7, draw=ocmain] (\sl,-9.00) rectangle (\sr,-9.64);
\node[icon] at (\ix,-9.32) {\includegraphics[height=14pt]{figures/icons/phosphor/robot.pdf}};
\node[txt] at (\tx,-9.32) {\textcolor{ocmain}{\textbf{verify \textperiodcentered{} retry \textperiodcentered{} switch}}\quad\textcolor{ocmuted}{action set intact}};

\end{tikzpicture}
\caption{A raw interface leaves detection, localization, and recovery
with the agent. Detect and advise instead flags a contract violation
and appends an advisory
receipt carrying a diagnostic witness and a list of available recovery
tools. It neither removes actions nor selects a repair. Controls
(Section~\ref{sec:results}) identify the recovery-tool list, not the
property-specific witness, as the active content in our tested sample.}
\label{fig:overview}
\end{figure}

%% file: sections/03-related-work.tex
\section{Related Work}

\paragraph{Tool-using language agents.}
ReAct interleaves natural-language reasoning with environment actions, while
Toolformer trains models to decide when and how to invoke
APIs~\cite{yao2023react,schick2023toolformer}. AgentBench and AppWorld
broaden evaluation to interactive multi-step
environments~\cite{liu2024agentbench,trivedi2024appworld}, subsequent
work scales tool repertoires~\cite{qin2024toolllm}, and SWE-agent shows
agent-computer interface design affects task
success~\cite{yang2024swe}; tool-hallucination work targets the agent's
own faulty call~\cite{zhang2024toolbehonest,xu2025reliability}.
These lines primarily evaluate planning and
tool selection under an assumed interface. We instead modify what the interface
communicates after a semantically suspect result.

\paragraph{Feedback and self-correction.}
Reflexion stores linguistic feedback across
trials~\cite{shinn2023reflexion}, CRITIC uses external tools to critique
and revise outputs~\cite{gou2024critic}, and Self-Refine iterates with
the same model as generator and critic~\cite{madaan2023selfrefine};
self-correction without external grounding can degrade
outputs~\cite{huang2023large}.
Reinforced Agent places a separate LLM reviewer before execution and
iteratively revises provisional tool calls~\cite{ta2026reinforced}; we
compare against a mechanism-faithful port in our experiments. Process supervision trains a
learned verifier to score each reasoning step~\cite{lightman2024lets};
our step-level receipts are instead deterministic, interface-derived
checks requiring no trained verifier and no extra model call.
PALADIN instead trains on failure-recovery trajectories and retrieves
recovery exemplars at inference time~\cite{vuddanti2026paladin}; ours is
training-free and targets implicit inconsistencies in returned results.
AgentProp-Bench reports little rank association between rejection and
recovery across nine models~\cite{gurram2026agentprop}, supporting the
distinction between detecting a bad interaction and recovering from it.

\paragraph{Runtime monitoring and verification.}
ToolEmu uses an LM-emulated sandbox to surface consequential agent
risks~\cite{ruan2024toolemu}. ToolMaze directly perturbs tool environments to
study dynamic replanning~\cite{zhu2026toolmaze}. $\tau$-bench evaluates agents
in stateful retail and airline domains~\cite{yao2024taubench}. Runtime
shields enforce safety constraints during
execution~\cite{alshiekh2018safe}. The closest recent systems make the
enforcement contrast concrete: ToolGate admits a tool call only when
Hoare-style pre/postconditions verify~\cite{liu2026toolgate}; Reason Less,
Verify More rejects policy-violating writes before
execution~\cite{reddy2026reason}; Agent Behavioral Contracts enforce
session-level governance rules~\cite{bhardwaj2026behavioral}; and further
systems supply rule languages, transactional compensation, or state-machine
postconditions with recovery owned by the
runtime~\cite{wang2026agentspec,chang2025sagallm,guo2026agentsama}.
ToolSafe trains a guardrail model to flag unsafe invocations before
execution~\cite{mou2026toolsafe}. Our monitor acts after a result arrives,
makes no enforcement decision, requires no trained verifier, and leaves
recovery with the acting agent. Post-trajectory false-success detectors
identify unsupported completion claims~\cite{advani2026false}; our checker
instead emits a per-call signal while the same trajectory can still recover.

\paragraph{Design by contract and runtime assertions.}
Checking a computed result against a declared expectation is a
long-standing software-engineering discipline: design by contract
specifies pre- and postconditions a component must
honor~\cite{meyer1992contract}, higher-order contract systems attribute
blame on violation~\cite{findler2002contractual}, and runtime
verification monitors execution against formal
properties~\cite{leucker2009runtime}. We port the interface-checking idea to language-agent
observations and explicitly choose an advisory propagation policy, so the
model, not the runtime, decides how to recover.

\paragraph{Learned invariants and specification mining.}
Because hand-written contracts are costly, a parallel line of work infers likely
specifications from observed executions. Daikon detects likely invariants from
dynamic traces~\cite{ernst2007daikon}, and specification-mining methods
generalize this to temporal and API-usage properties~\cite{ammons2002mining}.
We adopt the same non-oracle stance: our contracts are mined
from clean agent traces or derived from public response schemas rather than
authored against a task solution. Unlike program-analysis settings, the mined
invariant is not used to certify code but to explain a suspect observation to a
downstream agent at inference time.
Contract2Tool likewise infers preconditions and effects from documentation,
schemas, and traces, but deploys them for causal tool filtering
~\cite{babu2026contract2tool}; we cross-fit invariants over returned outcomes
and expose violations as advisory observations.

\paragraph{Agent safety and guardrails.}
Recent work proposes guardrail frameworks that filter or constrain agent
actions before or after execution~\cite{dong2024safeguarding,
inan2023llamaguard}. A related threat is adversarial payloads carried in
tool results themselves~\cite{zhan2024injecagent}; our threat model is instead
benign-but-inconsistent results; adversarial tools are out of scope.
Although nonbinding, false-positive context can still
induce the agent to take a harmful action.

\input{tables/alternatives}

Our experiments measure the advisory alternatives---generic caution,
always-verify, and a reviewer-loop port---under matched scaffolds; hard
enforcement remains a design contrast (Table~\ref{tab:alternatives}).

%% file: tables/alternatives.tex
% Shaded cells mark where a strategy differs from the Outcome Monitor row.
% The shading is descriptive, not evaluative: it locates design differences
% and does not rank them. Tint colour ocdiff is defined in main.tex.
\begin{table*}[t]
\centering
\small
\newcommand{\dif}[1]{\cellcolor{ocdiff}#1}
\begin{tabular}{llllll}
\toprule
Strategy & Trigger & Grounding & Localized witness & Restricts actions & Extra tool reads \\
\midrule
Generic caution & learned detector event & external invariant & \dif{no} & no & no \\
Self-critique & \dif{model's own judgment} & \dif{internal} & \dif{uncertain} & no & \dif{optional} \\
Always verify & \dif{every advertised read} & \dif{none (fixed policy)} & \dif{only after comparison} & no & \dif{yes} \\
Hard guard & monitor-detected event & external rule & yes & \dif{yes} & no \\
\midrule
\textbf{Outcome Monitor} & learned detector event & external invariant & yes & no & no \\
\bottomrule
\end{tabular}
\caption{Conceptual comparison of recovery interfaces. Shaded cells mark
where a strategy differs from the \textbf{Outcome Monitor} row, which
serves as the reference; the shading locates design differences and does
not rank them. The table characterizes mechanisms, not measured
superiority.}
\label{tab:alternatives}
\end{table*}

%% file: sections/04-methodology.tex
\section{Outcome Monitors}
\label{sec:method}

The system has two roles: a \emph{detector} that identifies when a tool result
violates a learned or public invariant, and an \emph{advisory signal} that
communicates the violation to the acting agent.
The mined invariants (outcome contracts) define the detector; the receipt
encodes the signal.
These components serve different roles: contracts determine when a result is
suspect, while public recovery relations determine which alternative tools the
receipt exposes. The controls isolate the receipt's active content; they do not
make contract-based detection dispensable.

\subsection{Interface}

At step $t$, an agent selects tool call $a_t$ from action set $A_t$ and receives
result $y_t$. A contract checker $C$ maps $(a_t,y_t)$ to an ordered set $V_t$
of violation records, each a machine-readable property code optionally
with a path, expected/observed values, and human-readable detail.
If $V_t$ is empty, the interface returns $y_t$ unchanged. Otherwise a
versioned encoder $E_s$ maps $(\tau,V_t,R_t)$ to receipt $\rho_t$, where $R_t$
contains only recovery affordances available from the public tool interface.
$A_t$ is unchanged. The agent may verify, retry, switch tools,
ignore the receipt, or take any other advertised action
(Figure~\ref{fig:overview}).

\begin{algorithm}[t]
\small
\caption{\textsc{CheckAndReceipt}: per-call contract evaluation.}
\label{alg:runtime}
\begin{algorithmic}[1]
\REQUIRE tool $\tau$, arguments $a$, result $y$, registry $C$, visible tools $T$, encoder $E_s$
\ENSURE augmented result; action set $A$ unchanged
\STATE $V \leftarrow C.\textsc{Check}(\tau, a, y)$
\IF{$V = \emptyset$}
  \RETURN $y$ \COMMENT{consistent}
\ENDIF
\STATE $R \leftarrow \{\tau\} \cup \{s \in \textsc{Subs}(\tau) : s \in T\}$
\STATE $\rho \leftarrow E_s(\tau,V,R)$ \COMMENT{versioned serialization}
\RETURN $\{\mathrm{tool\_result}:y,\;\mathrm{outcome\_contract}:\rho\}$
\end{algorithmic}
\end{algorithm}

Algorithm~\ref{alg:runtime} formalizes the per-call runtime: deterministic
given the call, result, and learned contract, and it never invokes tools,
scores candidate actions, or waits for task-level failure.

The frozen confirmation uses encoder $E_{v1}$; later controls use
$E_{v2}$ with an explicit witness structure. Algorithm~\ref{alg:runtime}
shows the ToolMaze envelope; in AppWorld and $\tau$-bench, whose tool
results are plain text, the receipt is appended as text (exact
envelopes in the supplement).

\subsection{Detector Construction}

The abstraction permits different non-oracle sources. In ToolMaze,
five-fold cross-fitting keeps every evaluated workflow out of the traces
used to build its registry. We mine those registries from nominal,
unperturbed (P0) trajectories using Algorithm~\ref{alg:learn}.

\begin{algorithm}[t]
\small
\caption{\textsc{LearnContracts}: invariant mining with acceptance rules.}
\label{alg:learn}
\begin{algorithmic}[1]
\REQUIRE nominal traces $D$, $K$ disjoint task folds $D_1,\ldots,D_K$
\ENSURE per-fold contract registry $C_1,\ldots,C_K$
\FOR{$k = 1$ \TO $K$}
  \STATE $D_{\mathrm{train}} \leftarrow D \setminus D_k$
  \FOR{each tool $\tau$ observed in $D_{\mathrm{train}}$}
    \STATE $F_{\mathrm{req}} \leftarrow \bigcap$ result field sets;\;
           $F_{\mathrm{all}} \leftarrow \bigcup$ result field sets
    \STATE $\mathrm{types}[f] \leftarrow$ union of kinds observed for $f$
    \STATE echo: accept $f$ iff $f$ is in $\ge 2$ call/result pairs
           and $a[f]\!=\!y[f]$ in all
    \STATE positivity: accept $f \in F_{\mathrm{req}}$ iff numeric in
           $\ge 3$ samples and always $>0$
    \STATE domain: accept $f \in F_{\mathrm{req}}$ iff name matches
           categorical lexicon, $\ge 4$ string samples, $\le 8$
           distinct normalized values
    \STATE affine: accept $(f,g) \in F_{\mathrm{req}}^2$ iff $\ge 4$
           samples with $\ge 3$ distinct values fit one
           $y[g]=\alpha\, y[f]+\beta$ ($10^{-8}$ tolerance)
    \STATE order: accept $(f,g)$ among three fixed date pairs iff
           $\ge 2$ parsed samples and $f \le g$ in all
  \ENDFOR
  \STATE $C_k \leftarrow$ registry over all accepted tool specs
\ENDFOR
\end{algorithmic}
\end{algorithm}

Acceptance is conservative. An invariant must hold on every training sample
and meet its class-specific support minimum (the categorical lexicon appears
in the supplement). At runtime, the checker evaluates each admitted class and
collects its violations in $V_t$; affine relations use a $10^{-5}$ tolerance
versus the tighter mining-time fit.

Recovery relations are constructed separately from detection. We build the
static mapping $\textsc{Subs}(\tau)$ programmatically from two public,
task-agnostic ToolMaze sources: per-tool substitute fields and the entry tool
of each sibling path in the alternatives file. The mapping therefore uses no
task descriptions, solutions, gold sequences, perturbations, or outcomes.
The indexing code and full mapping ship with the artifact. At runtime, we
intersect $\textsc{Subs}(\tau)$ with the workflow's advertised tools to form
$R_t$, so a receipt never names a tool the agent cannot call.
A task-disjoint audit of the detector on the experiments' difficult C3/C4
workflow classes flags
79.6\% of implicit faults at a 1.39\% clean false-positive rate, with a
learned contract available for 94.2\% of fault outcomes; rates are per
observed tool outcome, not per episode. An earlier
schema-only gate reached 55\% recall at an 18\% false-positive rate.
Section~\ref{sec:results} measures both detectors end-to-end, so the
audit characterizes detection precision rather than a completion
ceiling (full audit in the supplement).

For AppWorld, contracts derive from released standard API response schemas:
eligible read results are validated recursively, and a receipt fires on a
missing required field or a scalar type mismatch.

For $\tau$-bench retail, contracts are mined from the public database
(\texttt{orders.json}) without reading task descriptions, gold sequences, or
reward functions. The checker validates categorical domains (order status,
transaction type), positivity constraints (item prices), and affine
conservation relations (net payment equals item total for non-cancelled
orders). These are richer than schema-only contracts but still
non-oracle.

In all three environments, the injector changes only the text observed
by the model; environment state is never mutated, so evaluators score
the true state.

\paragraph{Construction cost.}
Construction is deterministic and local with zero model calls: the five
cross-fitted ToolMaze registries (15,041 invariants from 400 nominal
workflows), the AppWorld schema contracts, and the $\tau$-bench database
contracts each build in under two seconds (supplement).

\subsection{Information and Non-Oracle Boundary}

The runtime is prohibited from reading the task solution, perturbation label,
evaluator, gold API sequence, or hidden recovery path. Contract construction
uses only public tool metadata and disjoint clean traces. We audit the
remaining risk of a de facto oracle by
reporting the fraction of receipt events with at least two admissible public
recovery actions. This test is necessary but not sufficient: it shows the interface does
not usually reduce recovery to a single choice, not that every contract
is correct or every action useful.

%% file: sections/05-experimental-design.tex
\section{Experimental Design}

We ask three questions. \textbf{RQ1:} Does detector-triggered advisory
feedback improve full task completion under implicit tool failures? \textbf{RQ2:} How
does it compare to an inference-time reviewer alternative? \textbf{RQ3:} Does
the effect persist across environments and controls, and where does it fail?
Task identities, settings, and artifact hashes are frozen before the
corresponding outcomes are opened, except where noted; studies frozen or
amended after related outcomes were seen are labeled retrospective.

\subsection{ToolMaze Confirmation (RQ1)}

ToolMaze crosses workflow topology (C1--C4) with explicit/implicit and
transient/permanent tool perturbations~\cite{zhu2026toolmaze}. From the
difficult C3/C4 implicit-failure region we sample 80 unique workflow IDs,
exactly 20 each from C3/P3, C3/P4, C4/P3, and C4/P4 (transient and
persistent implicit faults), each carrying one fault mode.
For each model, baseline and advisory episodes are randomized and
interleaved, giving 160 episodes per model.

We evaluate DeepSeek V4 Flash and V4 Pro via the direct DeepSeek API and
Qwen 3.7 Plus and 3.7 Max via OpenRouter, all with identical prompts,
temperature zero, and extended reasoning disabled. The primary endpoint within
a model is paired full workflow completion, tested with the two-sided exact
McNemar/binomial test. Because all four models see the same 80 workflows,
joint inference clusters by workflow. We use an exact task-level sign-flip test
and a 200,000-resample task-cluster bootstrap.

A replication applies the unchanged 80-workflow method to MiniMax M3
through an exact MiniMax-hosted FP8 route on OpenRouter. It is analyzed
independently and is not pooled with the four-model primary.

\subsection{Method Comparison (RQ2)}

A retrospective study compares three arms on 80 additional ToolMaze
workflows: contemporaneous baseline, localized receipt, and a
mechanism-faithful Reinforced Agent
port~\cite{ta2026reinforced}. The port preserves separate pre-execution
review, generic v1 feedback, and up to five review loops, but substitutes
the matched DeepSeek tier for the source paper's GPT-4o base and reviewer;
it is therefore not an exact reproduction. Both DeepSeek tiers run all three
arms.

\subsection{Cross-Environment Studies (RQ3)}

\paragraph{AppWorld.}
AppWorld provides state-based tests that recognize alternative valid
solutions~\cite{trivedi2024appworld}. We deterministically select
one variant from each of the 19 generator families in the released development
split, and both DeepSeek tiers run fault baseline, fault advisory, and clean
advisory, with fault type alternating between a missing field and a scalar
type flip. A held-out study then selects 32 generator families
from \texttt{test\_normal} and runs the same three conditions on DeepSeek V4
Flash. The two samples are analyzed and reported separately.

\paragraph{$\tau$-bench retail.}
$\tau$-bench evaluates tool-calling agents in a stateful retail environment
with a simulated user and state-based reward~\cite{yao2024taubench}. A
factorial samples 50 of 115 retail test tasks with seed 20260718 and applies
both persistent read faults to every task: categorical status-out-of-set and
affine broken-conservation. Each task runs faulted baseline and advisory under
both faults plus clean advisory, yielding 250 episodes per tier at
temperature zero for each of DeepSeek V4 Flash and V4 Pro on the same
50 tasks. The
prespecified primary endpoint is paired completion within each fault class,
and the secondary aggregate is clustered by the 50 shared tasks. An earlier
exploratory study confounded fault class with task identity and suggested
a harmful conservation effect; the same-task design removes that confound.

\paragraph{Incident-derived faults.}
A two-tier study reruns the 80 confirmation workflows unperturbed
and injects one of eight fault types. We derive those types from the five
failure classes of a published taxonomy of production silent
failures~\cite{wu2026errors}, under a documented independence protocol:
the fault author saw only the taxonomy and fixture shapes, not the
contract classes. Each fault lands on the
tool the benchmark's own fault corrupts, and per-cell assignment is
frozen among the types expressible on that tool's nominal output. Every
injection's delivered result and field-level diff are logged. The
prespecified primary is intention-to-treat paired completion, and detection
is a receipt on the injected tool. Because a post-run audit found
imprecise placement, an earlier run was invalidated; the amended protocol
was frozen after that run's outcomes were known and is disclosed as such
(supplement).

\subsection{Controls}

Five control studies probe the receipt's active components, all using the
$E_{v2}$ encoder and paired workflow-clustered analyses. Three run on 57
reserved workflows: a \emph{specificity control} (generic warning vs.\
localized receipt), an \emph{always-verify control} (deterministic
substitute after each advertised read, frozen after specificity outcomes
and therefore retrospective), and a \emph{clean-P0 control} (no fault,
with the manifest frozen before any P0 outcome on these workflows was
opened). Two more run on 57 fresh workflows, with interpretations
registered before execution: a four-arm \emph{matched-trigger placebo}
varying only when a fixed generic warning attaches, and a three-arm
\emph{stripped-envelope} rerun that removes the recovery-tools list,
because a post-run audit found the always-warn envelope leaks affordances.

%% file: sections/06-results.tex
\section{Results}
\label{sec:results}

Prespecified, frozen tests are identified as such below; retrospective
and post hoc analyses are labeled where they appear. Inferential
statistics for the remaining control and boundary analyses are descriptive.

\subsection{Detector-Triggered Feedback Replicates Across Three Families}

\input{tables/toolmaze_main}

\textbf{The four-model primary is positive throughout.}
Table~\ref{tab:toolmaze-main} shows a positive, individually significant
effect for every model, and all 16 prespecified model-by-stratum point
estimates are nonnegative. Baseline succeeds on 35/320 model-workflow pairs
(10.9\%) and the advisory on 90/320 (28.1\%): a clustered mean effect of
$+17.2$~\pp{} (95\% bootstrap interval $[11.25,23.44]$, sign-flip
$p<.00001$). The task remains hard under the advisory, and 230/320 episodes
still fail.

\textbf{The separate third-family replication agrees.}
MiniMax M3 improves from 5/80 to 20/80 completions, $+18.75$~points with 17
paired wins and two losses (exact $p=.00073$). All four prespecified stratum
estimates are nonnegative.

All advisory wins contain at least one detector event. Of 484 events,
443 (91.5\%) retain at least two admissible recovery actions. After a
receipt, the agent switches to a listed substitute in 48.6\% of events,
calls another tool in 33.5\%, retries in 14.0\%, and makes no further
tool call in 3.9\%. The seven aggregate losses also contain receipts, so
localization does not guarantee replanning.

A reasoning ablation with 8,192-token thinking budgets shows effects
persist on both tiers (95\% interval $[10.00,25.62]$, combined sign-flip
$p<.0001$): neither standing fault awareness nor extended reasoning
substitutes for the delivered signal.
Both arms receive the same unmodified, fault-aware system prompt, which
already instructs the model to inspect unexpected tool outputs; no system
instruction mentions contracts or receipts. The intervention is entirely
in-band in the returned tool result.

\subsection{Benefits Replicate in a Stateful Retail Environment}

\input{figures/cross_environment_bars}

\textbf{A same-task factorial establishes a second positive environment.}
In the $\tau$-bench
factorial on DeepSeek V4 Flash, aggregate completion rises from 41/100 to
55/100: $+14.0$~points (95\% bootstrap interval $[3.0,25.0]$, task-clustered
sign-flip $p=.028$). Categorical status violations improve from 6/50 to
20/50, $+28.0$~points with 14/0 W/L (exact $p=.00012$). Broken-conservation
violations hold at 35/50 in both arms, an estimate of $0$~points, though the
9/9 W/L split shows pairs moving in both directions ($p=1.0$).
A matched DeepSeek V4 Pro run on the same 50 tasks confirms the
pattern: $+12.0$~points overall ($p=.023$), categorical $+18.0$~points
($p=.004$), conservation $+6.0$~points ($p=.607$). The conservation null is
consistent with any effect below the ${\approx}23$~points single-tier minimum
detectable effect at 80\% power; per-tier breakdowns and restriction
analyses appear in the supplement. Clean advisory emits one spurious receipt
across 100 episodes on the two tiers.

\subsection{Recovery Tools Carry the Detectable Receipt Gain}

\input{figures/signal_decomposition}

\paragraph{The recovery-tool list carries the detectable gain.}
Five matched contrasts vary one part of the receipt on the same 114
workflow pairs (Figure~\ref{fig:decomposition}). In the registered
three-arm rerun, the stripped receipt performs at baseline; restoring
the recovery-tool list adds $+11.4$~points ($p=.028$). Witness detail
and deferred timing produce no detectable differences, though both nulls
are power-bounded (${\approx}18$~points MDE). The always-warn comparison
is confounded because it supplies recovery tools at every step. Targeting
the same content with the detector reduces warning volume from
${\approx}600$ per tier to 80--91 without a detected completion loss
(per-tier estimates in the supplement).

\paragraph{Schema-only detector ablation.}
A contemporaneous three-arm study reruns the 80 confirmation cells on
DeepSeek V4 Flash. The learned arm replicates the confirmation
($13/80$ to $25/80$, $+15.0$~points, $p=.0075$). The
schema-only arm recovers much of the gain ($21/80$, $+10.0$~points, 11/3
W/L, $p=.057$); the head-to-head difference is $+5.0$~points (9/5 W/L,
$p=.42$). Detection behavior differs sharply: the schema arm fires 166
receipts to the learned arm's 111, with only 12.0\% listing multiple
recovery actions versus 91.9\%.
A companion clean-P0 study finds no measured completion harm from the
schema arm's extra false triggers ($p=.375$).

\paragraph{Clean behavior.}
On clean P0 workflows, advisory and baseline each complete 74/114 pairs,
an estimate of $0$~points (95\% interval $[-5.26,5.26]$).
The advisory emits 16 false-trigger receipts in 14 treated pairs, with
five paired rescues and five paired harms ($p=1.0$), so clean traffic
is not harm-free.

\subsection{Receipts Outperform the Evaluated Reviewer Port}

\textbf{Receipts outperform the reviewer port on these workflows, at a
fraction of its cost.}
In the retrospective 80-workflow comparison, full monitor receipts improve
over baseline by $+15.0$~points (95\% bootstrap interval $[6.875,23.75]$,
sign-flip $p=.0011$) and over a mechanism-faithful Reinforced Agent port by
$+11.25$~points ($[3.125,19.375]$, $p=.0144$); the reviewer improves over
baseline by $+3.75$~points ($[-2.5,10.0]$, $p=.349$). The receipt adds
\$0.13 in all-cache-miss list price ($+7\%$ of baseline cost) with no
additional model call; the reviewer adds \$9.24 ($+498\%$) through up to
five additional calls. The port is not an exact reproduction, so a calibrated
reviewer could be more accurate (per-tier counts in the supplement).

\subsection{Where Benefits and Detection Reach Their Limits}

\paragraph{Gains concentrate where faults block completion.}
Figure~\ref{fig:cross-environment} synthesizes the frozen studies alongside
two development rows. When faulted-baseline completion is at most 16\%
(ToolMaze 10.9\%, $\tau$-bench categorical 12.0\%), net gains are $+15.0$
to $+28.0$~points. When most baselines complete despite the fault
(conservation 70.0\%, held-out AppWorld 78.1\%), net effects are
$+6.3$~points or less with rescues offset by harms. Conditional rescue
rates are 21.8--34.6\% in seven of eight rows; $\tau$-bench Flash
conservation is the outlier (60\%, 9/15). This synthesis is post hoc and
descriptive, not a deployment criterion.

\paragraph{Held-out AppWorld: delivery without net gain.}
In the 19-family development characterization, type-flip faults produce
same-sign gains on both DeepSeek tiers, missing-field completion is
unchanged, and the schema checker emits zero receipts across 38 clean
episodes. In a held-out 32-family \texttt{test\_normal} study on DeepSeek
V4 Flash, baseline and advisory each complete 25/32 families ($p=1.0$).
The checker fires in all 32 faulted advisory episodes, but 23 pairs complete
despite the fault, leaving little room for rescue. Per-fault directions and
development details appear in the supplement.

\paragraph{Incident-derived faults expose the detection boundary.}
The suite transcribes fault content from real production incidents, authored
blind to the contract vocabulary. The unchanged detector flags 46.1\% and
45.3\% of injected faults on Flash and Pro. Recall is 83\% (25/30) for
violations expressible over structured values but 22\% (6/27) for corruption
inside plausible strings. Completion is null in both tiers ($p=1.0$), where
faulted baselines already complete 67.5\% and 65.0\% of episodes. Yet corrupted
values propagate in roughly one in five trackable episodes, and about two
thirds of those episodes still pass. The contracts' vocabulary therefore
bounds detection, while task completion can miss downstream corruption.

\paragraph{Organic traffic evaluates detection, not recovery.}
On 53,078 recorded StableToolBench responses from 427 public tools, with no
agent or counterfactual, the unchanged detector fires on 0.80\%, concentrated
in 54 tools, versus 1.39\% on synthetic clean
workflows~\cite{guo2024stabletoolbench}. Flags include HTTP errors inside
transport-success responses and empty payloads missing learned fields; not
every flag is adjudicated as a true failure. This measures detection, not recovery.

%% file: tables/toolmaze_main.tex
\begin{table}[t]
\centering
\small
\setlength{\tabcolsep}{3pt}
\begin{tabular}{@{}llrrrr@{}}
\toprule
Family & Model & Base & Monitor & $\Delta$ & W/L \\
\midrule
\icondeepseek\ DeepSeek & V4 Flash & 14/80 & 27/80 & +16.25 & 15/2 \\
\icondeepseek\ DeepSeek & V4 Pro & 13/80 & 23/80 & +12.50 & 14/4 \\
\iconqwen\ Qwen & 3.7 Plus & 3/80 & 15/80 & +15.00 & 12/0 \\
\iconqwen\ Qwen & 3.7 Max & 5/80 & 25/80 & +25.00 & 21/1 \\
\midrule
\multicolumn{2}{@{}l}{Primary aggregate} & 35/320 & 90/320 & +17.19 & 62/7 \\
\midrule
\iconminimax\ MiniMax & M3 replication & 5/80 & 20/80 & +18.75 & 17/2 \\
\bottomrule
\end{tabular}
\caption{Frozen paired ToolMaze completion. The four-model primary aggregate
clusters shared tasks; MiniMax M3 is a separately frozen third-family
replication and is not pooled.
$\Delta$ is in percentage points; W/L counts discordant pairs.
All models run with extended reasoning disabled.
Primary task-cluster sign-flip $p<.00001$;
95\% bootstrap interval $[11.25,23.44]$.}
\label{tab:toolmaze-main}
\end{table}

%% file: figures/cross_environment_bars.tex
% Paired dumbbell plot: baseline to advisory completion by study row.
% Open markers are baseline; filled markers are advisory. Coincident
% endpoints use a ring-and-dot glyph so zero changes remain visible.
% Counts in tab:cross-environment-supp. Drawn at print size; no resizebox.
\begin{figure}[t]
\centering
\begin{tikzpicture}[
  x=1cm, y=1cm,
  lbl/.style={font=\sffamily\scriptsize, text=ocink,
              anchor=west, inner sep=0pt},
  tag/.style={font=\sffamily\tiny\bfseries, text=ocmuted,
              anchor=west, inner xsep=2pt, inner ysep=0.6pt,
              rounded corners=1pt, fill=black!5},
  val/.style={font=\sffamily\scriptsize\bfseries,
              anchor=east, inner sep=0pt},
  tick/.style={font=\sffamily\scriptsize, text=ocmuted,
               anchor=north, inner sep=1pt},
  pair/.style={draw=ocmain!58, line width=1.4pt, line cap=round},
  base/.style={draw=ocink!72, fill=white, line width=0.8pt},
]
% x(v) = 3.25 + 0.037*v, v = completion percentage.
\def\ax{3.25}

% Column guides and axis.
\foreach \x in {4.175,5.10,6.025,6.95}{
  \draw[ocgrid, line width=0.4pt] (\x,0.10) -- (\x,-4.08);
}
\draw[ocink!65, line width=0.5pt] (\ax,-4.08) -- (6.95,-4.08);
\foreach \x/\t in {3.25/0,4.175/25,5.10/50,6.025/75,6.95/100}{
  \draw[ocink!65, line width=0.5pt] (\x,-4.08) -- (\x,-4.19);
  \node[tick] at (\x,-4.19) {\t};
}
\node[font=\sffamily\scriptsize, text=ocmuted, anchor=north]
  at (5.10,-4.48) {task completion (\%)};

% Direct key.
\draw[base] (3.35,0.28) circle (2.35pt);
\node[lbl] at (3.48,0.28) {baseline};
\fill[ocmain] (4.75,0.28) circle (2.35pt);
\node[lbl] at (4.88,0.28) {advisory};
\draw[base] (6.25,0.28) circle (2.65pt);
\fill[ocmain] (6.25,0.28) circle (1.35pt);
\node[lbl] at (6.38,0.28) {same};

% ToolMaze implicit, 10.9 -> 28.1.
\node[lbl] at (0,-0.42) {ToolMaze implicit (4 models)};
\draw[pair] (3.653,-0.42) -- (4.290,-0.42);
\draw[base] (3.653,-0.42) circle (2.35pt);
\fill[ocmain] (4.290,-0.42) circle (2.35pt);
\node[val, text=ocmain] at (8.30,-0.42) {$+$17.2};

% tau-bench status, 12.0 -> 40.0.
\node[lbl] at (0,-1.10) {$\tau$-bench status};
\draw[pair] (3.694,-1.10) -- (4.730,-1.10);
\draw[base] (3.694,-1.10) circle (2.35pt);
\fill[ocmain] (4.730,-1.10) circle (2.35pt);
\node[val, text=ocmain] at (8.30,-1.10) {$+$28.0};

% ToolMaze clean P0, 64.9 -> 64.9.
\node[lbl] at (0,-1.78) {ToolMaze clean P0};
\node[tag] at (2.18,-1.78) {CLEAN};
\draw[base] (5.651,-1.78) circle (2.65pt);
\fill[ocmain] (5.651,-1.78) circle (1.35pt);
\node[val, text=ocmuted] at (8.30,-1.78) {0.0};

% AppWorld type flip (development), 66.7 -> 83.3.
\node[lbl] at (0,-2.46) {AppWorld type flip};
\node[tag] at (2.10,-2.46) {DEV};
\draw[pair] (5.718,-2.46) -- (6.332,-2.46);
\draw[base] (5.718,-2.46) circle (2.35pt);
\fill[ocmain] (6.332,-2.46) circle (2.35pt);
\node[val, text=ocmain] at (8.30,-2.46) {$+$16.7};

% tau-bench conservation, 70.0 -> 70.0.
\node[lbl] at (0,-3.14) {$\tau$-bench conservation};
\draw[base] (5.840,-3.14) circle (2.65pt);
\fill[ocmain] (5.840,-3.14) circle (1.35pt);
\node[val, text=ocmuted] at (8.30,-3.14) {0.0};

% AppWorld missing field (development), 85.0 -> 85.0.
\node[lbl] at (0,-3.82) {AppWorld missing field};
\node[tag] at (2.62,-3.82) {DEV};
\draw[base] (6.395,-3.82) circle (2.65pt);
\fill[ocmain] (6.395,-3.82) circle (1.35pt);
\node[val, text=ocmuted] at (8.30,-3.82) {0.0};

\end{tikzpicture}
\caption{Paired baseline and advisory completion by study, sorted by
baseline rate. Open circles mark baseline and filled circles mark
advisory; concentric markers indicate identical rates. Development rows
are labeled \textsc{dev} and are not frozen tests. Rows are not pooled;
counts appear in the supplement.}
\label{fig:cross-environment}
\end{figure}
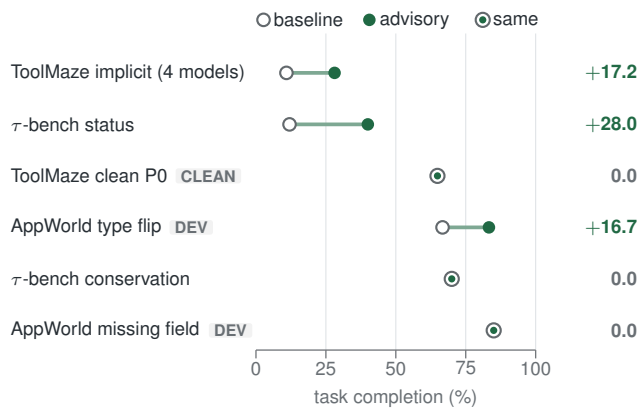

%% file: figures/signal_decomposition.tex
% Contrast plot: which receipt component moves completion.
% Five separately hashed contrasts, 114 workflow pairs each (57 per tier,
% two DeepSeek tiers). Estimates and intervals from tables
% tab:stripped-supp, tab:submission-controls-supp, tab:placebo-supp.
% Drawn at print size; no \resizebox. Label widths measured with \settowidth.
\begin{figure}[t]
\centering
\begin{tikzpicture}[
  x=0.99cm, y=1cm,
  lbl/.style={font=\sffamily\footnotesize, text=ocink,
              anchor=west, inner sep=0pt},
  grp/.style={font=\sffamily\scriptsize\bfseries,
              anchor=west, inner sep=0pt},
  val/.style={font=\sffamily\scriptsize\bfseries, anchor=east, inner sep=0pt},
  tick/.style={font=\sffamily\scriptsize, text=ocmuted,
               anchor=north, inner sep=1pt},
]
% x(v) = 4.603 + 0.1145*v   (v in percentage points)
\def\zero{4.603}

% Quiet group bands, then the zero reference and gridlines.
\fill[ocbluewash, rounded corners=2pt] (-0.03,0.16) rectangle (8.30,-1.04);
\fill[black!2, rounded corners=2pt] (-0.03,-1.18) rectangle (8.30,-2.78);
\draw[ocgrid, line width=0.4pt] (3.458,0.16) -- (3.458,-2.78);
\draw[ocgrid, line width=0.4pt] (5.748,0.16) -- (5.748,-2.78);
\draw[ocgrid, line width=0.4pt] (6.893,0.16) -- (6.893,-2.78);
\draw[ocmuted, line width=0.6pt, dash pattern=on 1.6pt off 1.4pt]
  (\zero,0.16) -- (\zero,-2.80);

% ---------------------------------------------------- group 1: list added
\node[grp, text=ocmain] at (0,0) {RECOVERY TOOLS ADDED};

\draw[ocmain, line width=1.1pt, line cap=round] (4.904,-0.42) -- (6.914,-0.42);
\fill[ocmain] (5.908,-0.42) circle (2.3pt);
\node[lbl] at (0,-0.42) {receipt $-$ stripped};
\node[val, ocmain] at (8.30,-0.42) {$+$11.4};

\draw[ocmain, line width=1.1pt, line cap=round] (5.005,-0.82) -- (7.114,-0.82);
\fill[ocmain] (6.009,-0.82) circle (2.3pt);
\node[lbl] at (0,-0.82) {receipt $-$ baseline};
\node[val, ocmain] at (8.30,-0.82) {$+$12.3};

% ------------------------------------------------ group 2: list unchanged
\node[grp, text=ocmuted] at (0,-1.34) {RECOVERY TOOLS UNCHANGED};

\draw[ocmuted, line width=1.1pt, line cap=round] (3.900,-1.76) -- (5.506,-1.76);
\draw[ocmuted, fill=white, line width=0.9pt] (4.704,-1.76) circle (2.3pt);
\node[lbl] at (0,-1.76) {stripped $-$ baseline};
\node[val, text=ocmuted] at (8.30,-1.76) {$+$0.9};

\draw[ocmuted, line width=1.1pt, line cap=round] (3.298,-2.16) -- (5.306,-2.16);
\draw[ocmuted, fill=white, line width=0.9pt] (4.302,-2.16) circle (2.3pt);
\node[lbl] at (0,-2.16) {localized $-$ generic};
\node[val, text=ocmuted] at (8.30,-2.16) {$-$2.6};

\draw[ocmuted, line width=1.1pt, line cap=round] (4.100,-2.56) -- (5.306,-2.56);
\draw[ocmuted, fill=white, line width=0.9pt] (4.704,-2.56) circle (2.3pt);
\node[lbl] at (0,-2.56) {on-time $-$ deferred};
\node[val, text=ocmuted] at (8.30,-2.56) {$+$0.9};

% ------------------------------------------------------------------ axis
\draw[ocink!65, line width=0.5pt] (3.000,-2.92) -- (7.350,-2.92);
\foreach \v/\t in {3.458/$-$10, 4.603/0, 5.748/10, 6.893/20}{
  \draw[ocink!65, line width=0.5pt] (\v,-2.92) -- (\v,-3.03);
  \node[tick] at (\v,-3.03) {\t};
}
\node[font=\sffamily\scriptsize, text=ocmuted, anchor=north] at (5.175,-3.34)
  {completion difference (points), 95\% interval};
\end{tikzpicture}
\caption{Contrast estimates with 95\% bootstrap intervals on 114 paired
workflows (57 per tier). Filled markers clear zero; open markers do not.
The three null contrasts have an ${\approx}18$~point minimum detectable
effect at 80\% power, bounding rather than establishing equivalence.}
\label{fig:decomposition}
\end{figure}
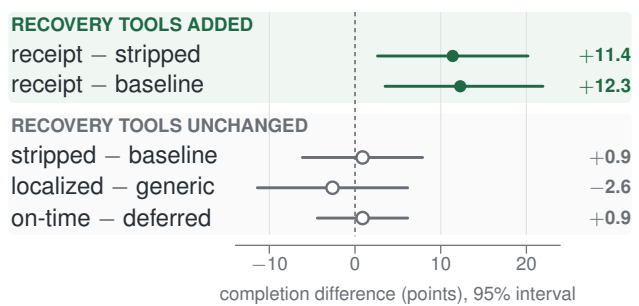

%% file: sections/07-discussion.tex
\section{Discussion and Limitations}

\paragraph{Why nonbinding matters.}
Because the receipt is nonbinding, a false positive can only add context
and salience; every action stays available. That does not eliminate
harm: the paired losses show an agent can overreact to correct but
irrelevant information, so preserving the action space is not a formal
safety shield.

\paragraph{Deployment value depends on fault prevalence.}
Every positive study uses a fault-enriched sample, so we bound the
deployment mixture by fault prevalence $\pi$. At the frozen point
estimates the advisory is net-nonnegative at every prevalence; at the
unfavorable interval endpoints of both studies a positive net effect
requires $\pi\gtrsim 0.32$ (supplement). At low prevalence the case
rests on the clean effect staying near its observed zero. A deployment
should therefore audit both its clean receipt rate and fault prevalence
before relying on the observed net benefit.

\paragraph{What the signal contributes.}
The controls point to recovery tools as the active content in these
ToolMaze samples; null contrasts for witness detail, timing, and salience
remain power-bounded. The schema ablation suggests that learned contracts
buy selectivity and better recovery lists rather than a clearly larger
completion effect.

\paragraph{Detector--injector alignment.}
We designed the $\tau$-bench and AppWorld faults so that their contract
sources could express them, so those studies test whether the signal is
actionable, not whether the detector finds arbitrary faults. ToolMaze
draws on the benchmark's own fault taxonomy and on generic
Daikon-lineage properties, but the miner and the injector may still be
aligned. The incident-derived suite addresses that: on faults authored
blind to the vocabulary, detection falls to ${\approx}46\%$, high on
violations a value can express and low inside plausible strings. No
study here measures recovery from failures that occur naturally.

\paragraph{Limitations.}
No episode here runs against a live deployment. What we inject is
placement and prevalence rather than fault content: the incident suite's
eight types come from a published taxonomy of production silent
failures~\cite{wu2026errors}, authored blind to the contract vocabulary,
and the unchanged detector also runs on 53,078 recorded public-API
responses. Recovery from organic failures remains open.
ToolMaze spans five models
from three provider families, but $\tau$-bench covers one family and the signal
decomposition only two DeepSeek tiers. Contracts inherit their sources'
blind spots and false positives: schema contracts cover fewer modes, and
AppWorld's development sample leans on two domains. Completion is a blunt
measure too. Clean controls record paired harms even where the
aggregate effect is zero, and an episode that propagates a corrupted
value can still pass. Nothing here establishes safety under adversarial
tools~\cite{zhan2024injecagent}, privacy failures, or irreversible
actions.

\paragraph{Reproducibility.}
The supplement and anonymous artifact report the full 4,683-episode study
inventory together with sanitized manifests, aggregate analyses, code, hashes,
and invalidation reports. Row-level provider and protected benchmark
observations are excluded from the review package.

%% file: sections/08-conclusion.tex
\section{Conclusion}

\method{} detect contract violations and expose recovery tools without
restricting the agent. Gains replicate across three provider families and
two environments but concentrate where faults block completion. The results
frame silent-failure recovery as an observability problem: viable actions may
already exist, but the interface must make them legible. Detection beyond the
contract vocabulary remains open.

%% file: sections/supplement.tex
This supplement accompanies the main paper and provides additional tables,
analysis details, and reproducibility information. The main paper is
self-contained; this material supports deeper inspection.

\section{Reasoning Ablation}

\begin{supptable}
\centering
\small
\begin{tabular}{llrrrr}
\toprule
Model & Think & Base & Contract & $\Delta$ & W/L \\
\midrule
V4 Flash & off & 14/80 & 27/80 & +16.25 & 15/2 \\
V4 Flash & 8k & 13/80 & 25/80 & +15.00 & 15/3 \\
V4 Pro & off & 13/80 & 23/80 & +12.50 & 14/4 \\
V4 Pro & 8k & 14/80 & 30/80 & +20.00 & 16/0 \\
\midrule
\multicolumn{2}{l}{Think-on combined} & 27/160 & 55/160 & +17.50 & 31/3 \\
\bottomrule
\end{tabular}
\caption{Contract effects persist in every configuration. Reasoning ablation on
DeepSeek V4 with an 8,192-token thinking budget.
Combined thinking-on task-cluster sign-flip $p<.0001$;
95\% bootstrap interval $[10.00,25.62]$.}
\label{tab:reasoning-ablation-supp}
\end{supptable}

\section{Paired Outcome Decomposition}

\begin{supptable}
\centering
\scriptsize
\setlength{\tabcolsep}{2pt}
\begin{tabular}{@{}llrrrr@{}}
\toprule
Environment & Model & Rescue & Harm & Both pass & Both fail \\
\midrule
ToolMaze & DS Flash & 15 & 2 & 12 & 51 \\
ToolMaze & DS Pro & 14 & 4 & 9 & 53 \\
ToolMaze & Qw Plus & 12 & 0 & 3 & 65 \\
ToolMaze & Qw Max & 21 & 1 & 4 & 54 \\
ToolMaze repl. & MiniMax M3 & 17 & 2 & 3 & 58 \\
  AppWorld dev & DS Flash & 3 & 1 & 13 & 2 \\
  AppWorld dev & DS Pro & 2 & 1 & 14 & 2 \\
  AppWorld test & DS Flash & 2 & 2 & 23 & 5 \\
\bottomrule
\end{tabular}
\caption{Paired outcome decomposition. Rescue means advisory-only success;
harm means baseline-only success. AppWorld development and held-out test
samples remain separate. MiniMax M3 is a separately frozen replication and is
not pooled with the four-model primary.}
\label{tab:pair-decomposition-supp}
\end{supptable}

\section{Post-Receipt Recovery Behavior}

\begin{supptable}
\centering
\small
\begin{tabular}{lrr}
\toprule
First action after receipt & Events & Share \\
\midrule
Listed recovery substitute & 235 & 48.6\% \\
Different tool & 162 & 33.5\% \\
Retry receipted tool & 68 & 14.0\% \\
No further tool call & 19 & 3.9\% \\
\bottomrule
\end{tabular}
\caption{Descriptive first-action distribution after 484 receipt events in
the four frozen ToolMaze confirmations. The categories characterize retained
recovery behavior; they do not identify an effect of receipt wording.}
\label{tab:post-receipt-supp}
\end{supptable}

\section{Reviewer-Port Comparison Counts}

\begin{supptable}
\centering
\small
\begin{tabular}{lrrr}
\toprule
Strategy & Flash & Pro & List cost \\
\midrule
Baseline & 16/80 & 13/80 & \$1.853 \\
Localized receipt & 27/80 & 26/80 & \$1.984 \\
Reinforced Agent port & 19/80 & 16/80 & \$11.089 \\
\bottomrule
\end{tabular}
\caption{Matched ToolMaze comparison on 80 separately hashed workflows
(main paper, Results). Counts are completed workflows per tier; cost is the
two-tier all-cache-miss list price. The port substitutes matched DeepSeek
tiers for the source paper's GPT-4o and is not an exact reproduction.}
\label{tab:published-comparison-supp}
\end{supptable}

\section{Detector Development and Audit}

\begin{supptable}
\centering
\small
\begin{tabular}{lrrr}
\toprule
ToolMaze detector & Impl. recall & Clean FPR & Coverage \\
\midrule
Schema-only gate & 55.0\% & 18.0\% & --- \\
Learned C2 gate & 80.0\% & 4.0\% & --- \\
C3/C4 cross-fit audit & 79.6\% & 1.39\% & 94.2\% \\
\bottomrule
\end{tabular}
\caption{The initial schema-only gate failed its pre-model audit. Outcome-level
detector development and task-disjoint audit. Coverage is the share of
implicit fault outcomes with a learned contract available; recall given
coverage is 84.5\% in the C3/C4 audit.}
\label{tab:detector-audit-supp}
\end{supptable}

The audit rate is per observed outcome rather than per episode, so
repeated opportunities accumulate. In the separate clean-P0 control they
reach 16 false contract events in 14/114 treated episodes, with five
paired harms and five paired rescues.

\section{Schema-Only Detector Ablation}

\begin{supptable}
\centering
\small
\setlength{\tabcolsep}{3pt}
\begin{tabular}{lrrrr}
\toprule
Contrast & Completion & $\Delta$ & W/L & $p$ \\
\midrule
Learned vs.\ baseline & 13/80 $\to$ 25/80 & +15.0 & 15/3 & .0075 \\
Schema-only vs.\ baseline & 13/80 $\to$ 21/80 & +10.0 & 11/3 & .0574 \\
Learned vs.\ schema-only & 21/80 $\to$ 25/80 & +5.0 & 9/5 & .4240 \\
\bottomrule
\end{tabular}
\caption{Frozen three-arm schema ablation (protocol
OUTCOME\_CONTRACT\_SCHEMA\_ABLATION\_80\_V1, manifest hash 592d5a43) on the
exact 80 frozen confirmation cells, DeepSeek V4 Flash, all three arms
contemporaneous, temperature 0. The learned arm contemporaneously replicates
the frozen confirmation effect. Mechanism: the learned arm emits 111 receipt
events on 67 cells (91.9\% listing $\ge 2$ admissible recovery actions); the
schema-only arm emits 166 events on 65 cells (12.0\%), because the public
registry carries no recovery-start relations. Every cell is faulted, so this
sample does not exercise the schema detector's 18\% clean false-positive
rate.}
\label{tab:schema-ablation-supp}
\end{supptable}

\section{Incident-Derived Fault Suite}
\label{sec:incident-supp}

\paragraph{Derivation and independence protocol.}
Eight fault types were transcribed from the five-class production-incident
taxonomy of Wu (2026, arXiv:2606.14589; 22 public postmortems). The fault
author was shown only the taxonomy section of that paper, two P0 fixture
files, and the transformation API contract, never the contract classes, learned
registries, or this paper. They returned deterministic transformation
algorithms with per-fault incident provenance. The
returned specification was implemented verbatim (two internal
example/algorithm inconsistencies were resolved in favor of the
algorithm text and logged). Per cell, the fault is injected on the first eligible result from the
exact tool(s) the benchmark's own P3/P4 fault corrupts. Assignment is
frozen among the fault types that change the target tool's nominal
fixture output, recomputed by the manifest validator from hashed
fixtures. Every injection records the delivered result and a field-level
before/after diff.

\paragraph{Invalidated first run (disclosure).}
A first run of this suite used step-count placement. A post-run audit
then found that only 58\% of its injections hit the benchmark-targeted
tool, that its detection count included receipts on non-injected tools,
and that delivered content was not recorded. The run was therefore
invalidated, and its numbers are not reported. The amended protocol (tool-level targeting, same-tool
detection, delivered-content audit trail, eligibility-aware assignment)
was frozen \emph{after} that run's outcomes were known.

\begin{supptable}
\centering
\footnotesize
\setlength{\tabcolsep}{2pt}
\begin{tabular}{lrr}
\toprule
Endpoint & Flash & Pro \\
\midrule
Primary ITT (base $\to$ adv.) & 54/80 $\to$ 54/80 & 52/80 $\to$ 52/80 \\
\quad paired W/L, exact $p$ & 3/3, $p=1.0$ & 4/4, $p=1.0$ \\
Injected episodes & 152/160 & 149/160 \\
Same-tool detection & 35/76 (46.1\%) & 34/75 (45.3\%) \\
Coincidental-receipt episodes & 11 & 10 \\
Contaminated / trackable (base) & 12/61 & 15/61 \\
\quad of which passed & 7 & 10 \\
\bottomrule
\end{tabular}
\caption{Incident-derived fault suite, protocol
OUTCOME\_\allowbreak CONTRACT\_\allowbreak INCIDENT\_\allowbreak
FAULT\_\allowbreak 80\_\allowbreak V2 (Flash manifest f4e9924a, Pro
manifest d39bd825), 160 episodes per tier on the 80 frozen confirmation
workflows with bit-identical per-task fault assignment across tiers.
Contamination is a string-match lower bound: a distinctive injected value
($\ge 8$ characters) appearing in the trace demonstrates propagation into
a later real tool call or the final answer, because the engine logs only
pre-injection results. Non-injections are attributed: target tool never called (3 episodes per
arm on Flash; 5 baseline and 4 advisory episodes on Pro) or live target
result ineligible for the assigned transform (1 episode per arm, both
tiers).}
\label{tab:incident-supp}
\end{supptable}

\begin{supptable}
\centering
\small
\setlength{\tabcolsep}{3pt}
\begin{tabular}{llrr}
\toprule
Class & Fault type & Det.\ (Flash) & Det.\ (Pro) \\
\midrule
C & vacuous\_exec\_swallow & 10/10 & 8/8 \\
B & positional\_field\_shift & 8/10 & 8/10 \\
E & stale\_declared\_state & 7/10 & 7/10 \\
C & cause\_stripped\_http\_payload & 3/9 & 4/10 \\
C & error\_page\_as\_payload & 3/10 & 3/10 \\
D & context\_pollution\_fabrication & 3/10 & 3/10 \\
B & ref\_id\_substitution & 1/9 & 1/9 \\
A & env\_threshold\_truncation & 0/8 & 0/8 \\
\bottomrule
\end{tabular}
\caption{Same-tool detection by fault type (denominators are changed
injections). Classes follow Wu (2026): A environment quirks,
B design-assumption mismatches, C error swallowing/dilution, D chained
fabrication, E operational omission. The gradient separates value-level
violations (caught by domain, positivity, and echo invariants) from
content-level corruption inside plausible strings (missed).}
\label{tab:incident-by-fault-supp}
\end{supptable}

\section{Cross-Environment Counts}

\begin{supptable}
\centering
\scriptsize
\setlength{\tabcolsep}{2pt}
\begin{tabular}{@{}llrrrl@{}}
\toprule
Env. & Fault class & Base & Adv. & $\Delta$ & Status \\
\midrule
ToolMaze & implicit (4 models) & 35/320 & 90/320 & +17.2 & frozen \\
$\tau$-bench & status (Flash) & 6/50 & 20/50 & +28.0 & frozen \\
$\tau$-bench & conserv. (Flash) & 35/50 & 35/50 & 0.0 & frozen \\
AppWorld dev & type flip & 12/18 & 15/18 & +16.7 & dev \\
AppWorld dev & missing field & 17/20 & 17/20 & 0.0 & dev \\
\bottomrule
\end{tabular}
\caption{Underlying counts for the main paper's descriptive synthesis. Rows
are not pooled. The figure's 95\% intervals are task-cluster bootstrap
intervals from the same run ledgers:
$\tau$-bench status $[16.0,40.0]$, conservation $[-16.0,16.0]$;
AppWorld type flip $[0.0,33.3]$, missing field $[-20.0,20.0]$;
ToolMaze uses the frozen confirmation interval $[11.25,23.44]$. The
$\tau$-bench intervals resample the 50 tasks shared by both fault classes.}
\label{tab:cross-environment-supp}
\end{supptable}

\section{AppWorld Studies}

The development characterization and held-out confirmation are distinct
studies. The former covers 19 generator families on both DeepSeek tiers and is
used for fault-class analysis. The latter freezes 32 families from
\texttt{test\_normal} and evaluates DeepSeek V4 Flash only. Its manifest listed
both DeepSeek tiers as planned, but no Pro episodes were executed or analyzed.

\begin{supptable}
\centering
\small
\setlength{\tabcolsep}{3pt}
\begin{tabular}{lrrrrr}
\toprule
Sample & Base & Adv. & W/L & Clean & Inj. \\
\midrule
Dev, Flash & 14/19 & 16/19 & 3/1 & 15/19 & 100\% \\
Dev, Pro & 15/19 & 16/19 & 2/1 & 16/19 & 100\% \\
Held-out test, Flash & 25/32 & 25/32 & 2/2 & 26/32 & 100\% \\
\bottomrule
\end{tabular}
\caption{The held-out \texttt{test\_normal} study has zero net effect (McNemar
$p=1.0$) despite fault injection in all 64 faulted episodes. AppWorld
studies are reported separately. Injection is the fraction of
faulted episodes receiving the protocol-specified corruption.}
\label{tab:appworld-studies-supp}
\end{supptable}

\subsection{Fault-Type Breakdowns}

\begin{supptable}
\centering
\scriptsize
\setlength{\tabcolsep}{2pt}
\begin{tabular}{lllrrr}
\toprule
Sample & Fault & Model & Base & Contract & $\Delta$ \\
\midrule
Dev & Missing field & Flash & 8/10 & 8/10 & 0 \\
Dev & Missing field & Pro & 9/10 & 9/10 & 0 \\
Dev & Type flip & Flash & 6/9 & 8/9 & +22.2 \\
Dev & Type flip & Pro & 6/9 & 7/9 & +11.1 \\
Held-out & Missing field & Flash & 12/16 & 14/16 & +12.5 \\
Held-out & Type flip & Flash & 13/16 & 11/16 & $-$12.5 \\
\bottomrule
\end{tabular}
\caption{The development by-fault direction does not replicate in the held-out
sample. AppWorld completion by fault assignment, with development and
held-out samples kept separate.}
\label{tab:appworld-fault-supp}
\end{supptable}

\section{$\tau$-bench Detailed Results}

The frozen factorial samples 50 retail test tasks and applies both
persistent \texttt{get\_order\_details} faults to every task, so each
task contributes two baseline/advisory pairs plus one clean-advisory
episode. The 250-episode DeepSeek V4 Flash run completes without errors:
all 200 faulted episodes fire the intended corruption, and none of the
50 clean episodes emits a receipt.

\begin{supptable}
\centering
\small
\begin{tabular}{lrrrr}
\toprule
Fault class & Base & Contract & $\Delta$ & W/L \\
\midrule
Categorical status & 6/50 & 20/50 & +28.0 & 14/0 \\
Conservation & 35/50 & 35/50 & 0.0 & 9/9 \\
\midrule
Aggregate & 41/100 & 55/100 & +14.0 & 23/9 \\
\bottomrule
\end{tabular}
\caption{Frozen same-task $\tau$-bench factorial. The categorical primary
comparison has exact paired $p=.00012$. The secondary aggregate clusters both
fault contrasts by task ($p=.028$; 95\% bootstrap interval $[3.0,25.0]$).}
\label{tab:taubench-factorial-supp}
\end{supptable}

An earlier 115-task-per-tier exploratory run assigned fault class by task-index
parity. Its harmful conservation estimate is retained as audit history but does
not persist in the deconfounded factorial design.

\section{Control Study Counts and Power Analysis}

\begin{supptable}
\centering
\small
\begin{tabular}{lrrr}
\toprule
Control & Flash & Pro & Joint $\Delta$ [95\% CI] \\
\midrule
Always verify & 11$\to$12 & 16$\to$18 & +2.63 [$-2.63$,7.89] \\
Clean P0 advisory & 38$\to$40 & 36$\to$34 & 0.00 [$-5.26$,5.26] \\
\bottomrule
\end{tabular}
\caption{Separately hashed ToolMaze controls, each with 57 workflows per
tier. Counts are baseline$\to$treatment completions; joint $\Delta$ and its
interval are in percentage points; intervals cluster the shared tasks.
Always verify is retrospective and has only 19 verification reads. The
specificity control's joint counts (generic caution 43/114, localized
40/114, $\Delta=-2.63$ [$-11.40$,6.14]) appear in the main text.}
\label{tab:submission-controls-supp}
\end{supptable}

The control studies (specificity, always-verify, clean-P0, $\tau$-bench
conservation) all report null results, so what matters is how much each
null can rule out. The minimum detectable effect (MDE) at 80\% power
quantifies that.

\begin{supptable}
\centering
\small
\setlength{\tabcolsep}{3pt}
\begin{tabular}{lrrr}
\toprule
Control study & $n$ pairs & Observed $\Delta$ & MDE \\
\midrule
Specificity (localized vs.\ generic) & 114 & $-$2.63 pp & 18 pp \\
Always-verify & 114 & +2.63 pp & 19 pp \\
Clean-P0 & 114 & 0.0 pp & 17 pp \\
Placebo timing (true vs.\ placebo) & 114 & +0.88 pp & 18 pp \\
Salience (stripped vs.\ baseline) & 114 & +0.88 pp & 18 pp \\
$\tau$-bench conservation & 50 & 0.0 pp & 23 pp \\
\bottomrule
\end{tabular}
\caption{Simulation-based MDE (sign-flip test, 10{,}000 simulations per
delta, $\alpha=.05$, seed 42). All controls are underpowered to detect
effects below ${\approx}\,17$--$23$ \pp{}; null results cannot be
interpreted as equivalence. The timing and salience contrasts of the
decomposition studies below share the specificity design (114 pairs,
sign-flip) and therefore its ${\approx}18$~\pp{} MDE.}
\label{tab:power-analysis-supp}
\end{supptable}

\section{Warning-Signal Decomposition Controls}
\label{sec:decomposition-supp}

\paragraph{Matched-trigger placebo.}
Protocol OUTCOME\_\allowbreak CONTRACT\_\allowbreak MATCHED\_\allowbreak
PLACEBO\_\allowbreak 57\_\allowbreak V1 samples 57 fresh C3/C4 workflows
from the task IDs appearing in no prior manifest (strata c3/P3 14, c3/P4
15, c4/P3 14, c4/P4 14) and runs four arms per tier (228 episodes per
tier). All warned arms deliver the byte-identical generic caution of the
specificity control inside the same envelope; the arms differ only in
when the warning attaches. The matched placebo suppresses the warning at
each detected violation, forwards the violating result raw, and delivers
one queued identical warning at the nearest following detector-consistent
result, the only online-implementable position match. Of the suppressed
violations, 75 of 77 (Flash) and 80 of 82 (Pro) had their warning
delivered, and any undelivered pending warning is recorded.

\begin{supptable}
\centering
\small
\begin{tabular}{lrr}
\toprule
Arm & Flash & Pro \\
\midrule
Baseline & 12/57 & 14/57 \\
True trigger & 16/57 & 17/57 \\
Matched placebo & 17/57 & 15/57 \\
Always warn & 20/57 & 26/57 \\
\bottomrule
\end{tabular}
\caption{Matched-trigger placebo completions. Registered primary
(true trigger minus matched placebo, two-tier workflow clusters):
$+0.88$~\pp, sign-flip $p=1.0$, 95\% bootstrap interval $[-4.39,6.14]$;
per-tier W/L 3/4 (Flash, $p=1.0$) and 3/1 (Pro, $p=.625$). Secondary
arm-minus-baseline estimates: true trigger $+7.0$ ($p=.344$) and $+5.3$
($p=.453$); matched placebo $+8.8$ ($p=.180$) and $+1.8$ ($p=1.0$);
always warn $+14.0$ ($p=.039$) and $+21.1$ ($p=.004$) for Flash and Pro,
respectively.}
\label{tab:placebo-supp}
\end{supptable}

\paragraph{Always-warn confound (audit disclosure).}
The registered envelope attaches
\texttt{admissible\_recovery\_tools} to every warned result. A post-run
audit found that no other channel exposes substitute names (0 of 270 tool
descriptions) and that on Pro nine of the fourteen always-warn paired
wins first call an envelope-surfaced substitute. Excluding those wins moves
the always-warn-versus-baseline $p$ from $.004$ to $.45$ (Flash: $.039$
to $.065$). The exclusion was not registered and is diagnostic
only; the stripped-envelope study tests the hypothesis prospectively. The
detector is additionally silent in 9 of 57 workflows, where only a
constant arm delivers any warning, so part of the always-warn margin is
recall coverage rather than timing.

\paragraph{Stripped-envelope study.}
Protocol OUTCOME\_\allowbreak CONTRACT\_\allowbreak STRIPPED\_\allowbreak
ALWAYS\_\allowbreak WARN\_\allowbreak 57\_\allowbreak V1 reruns the same
57 cells contemporaneously with three arms per tier: baseline, the full
always-warn envelope, and the identical envelope with the
\texttt{admissible\_recovery\_tools} key absent. The registered primary
is always-warn-stripped minus baseline; the registered secondaries are
full minus stripped (the leakage estimate) and full minus baseline (the
contemporaneous replication). Interpretations for each outcome were
frozen before execution.

\begin{supptable}
\centering
\small
\begin{tabular}{lrr}
\toprule
Arm & Flash & Pro \\
\midrule
Baseline & 17/57 & 12/57 \\
Always warn (full) & 19/57 & 24/57 \\
Always warn (stripped) & 14/57 & 16/57 \\
\bottomrule
\end{tabular}
\caption{Stripped-envelope study completions. Two-tier joint contrasts:
stripped minus baseline $+0.88$~\pp{} ($p=1.0$, $[-6.14,7.89]$); full
minus stripped $+11.40$ ($p=.028$, $[2.63,20.18]$); full minus baseline
$+12.28$ ($p=.018$, $[3.51,21.93]$). Per-tier full minus stripped: Flash
$+8.8$ (9/4, $p=.267$), Pro $+14.0$ (10/2, $p=.039$). Pro reproduces its
frozen always-warn estimate ($12\to24$, 13/1, $p=.0018$); Flash does not
($17\to19$, $+3.5$, $p=.774$), its contemporaneous baseline sitting five
completions above the frozen value with a stable always-warn arm.}
\label{tab:stripped-supp}
\end{supptable}

Warned arms fire on every tool result (617/599 full and 550/540 stripped
events for Flash/Pro, against 80--91 for the triggered arm of the placebo study). Envelope
integrity is verified per event in both ledgers: the full arm carries
the affordance key in every event and the stripped arm in none. Both
tiers complete with zero infrastructure retries.

\section{Per-Corruption-Type Detector Recall}

The cross-fitted audit covers nine corruption types across the C3/C4
region, and recall varies substantially by fault class.

\begin{supptable}
\centering
\small
\begin{tabular}{lrrr}
\toprule
Corruption type & $n$ faults & Recall & Coverage \\
\midrule
Logic violation & 408 & 91.2\% & 97.5\% \\
Physics violation & 463 & 85.3\% & 93.5\% \\
Temporal contradiction & 393 & 82.4\% & 95.9\% \\
Missing required fields & 328 & 80.2\% & 95.7\% \\
Tool leakage & 347 & 79.8\% & 87.3\% \\
Internal contradiction & 195 & 77.9\% & 94.9\% \\
Fact contradiction & 596 & 76.2\% & 92.3\% \\
Irrelevant response & 323 & 70.3\% & 96.6\% \\
Statistical inconsistency & 354 & 70.1\% & 94.9\% \\
\bottomrule
\end{tabular}
\caption{Per-corruption-type detector recall (C3/C4 cross-fitted audit,
sorted by recall). Coverage is the share of fault outcomes with a learned
contract available; recall given coverage exceeds 73\% for all types.}
\label{tab:per-corruption-supp}
\end{supptable}

\section{Prompts and Receipt Texts}

This section reproduces the exact texts the models received. Unicode em
dashes and one emoji glyph in the original source strings are rendered
below as \texttt{--} and \texttt{[emoji]}; everything else is verbatim.

\subsection{ToolMaze agent system prompt}

Baseline and advisory arms use the identical, unmodified fault-aware
system prompt from the pinned ToolMaze checkout
(\texttt{evaluation/agents/openai\_agent.py}), shown in
Figure~\ref{box:system-prompt}. No arm receives any
instruction mentioning contracts, receipts, or the
\texttt{outcome\_contract} field: the receipt is purely in-band.

\begin{figure*}[t]
\begin{promptbox}[top=3pt,bottom=3pt,
before upper={\scriptsize\ttfamily\raggedright}]{ToolMaze agent system prompt (P1--P4, identical in baseline and advisory arms)}
You are an autonomous AI assistant that executes tool-based workflows.
Tools may occasionally fail or return unexpected data.\\[3pt]
TASK COMPLETION\\
- The user query may describe multiple sub-tasks. You must complete ALL
of them -- do NOT skip any sub-tasks.\\
- Every fact, data retrieval, or state change must come from a tool
call. Never substitute a tool result with internal knowledge, and never
report a step as completed unless you actually called the corresponding
tool and received its result in this conversation.\\
- Sub-tasks may be accomplishable through different tool combinations.\\[3pt]
FAULT HANDLING\\
- When evaluating tool outputs, consider whether the returned data is
coherent and usable for the next step -- fields should be present,
values should be semantically reasonable, and types should match what
the workflow requires.\\
- If you determine that a required step cannot be accomplished by any
available tool or approach, stop the workflow and clearly report which
step failed and why, rather than proceeding with incomplete results.\\[3pt]
FINAL ANSWER FORMAT\\
- Be CONCISE and DIRECT. No emojis, markdown formatting (**, [emoji],
etc.), or bullet points.\\
- If SUCCESSFUL: State only essential facts and requested results
(e.g., confirmation codes). Omit internal technical details.\\
- If HALTED: State clearly that the task was aborted and the reason
(e.g., ``Task aborted: the pricing tool returned corrupted data after
retry and no other alternative approaches exist''). Do not apologize
excessively.
\end{promptbox}
\vspace{-4pt}
\caption{Verbatim ToolMaze system prompt.}
\label{box:system-prompt}
\end{figure*}

\subsection{ToolMaze receipt envelope}

On a detected violation the interface replaces the tool message content
with the JSON serialization of
\texttt{\{"tool\_result": <raw result>, "outcome\_contract": <receipt>\}};
the raw result is always preserved. A real receipt from the frozen
confirmation ledger (schema v1; task with the corrupted stock quote):

\begin{jsonbox}{Receipt, schema v1 (frozen confirmation)}
{
  "status": "inconsistent",
  "violations": [
    {"code": "nonpositive_value",
     "detail": "nominally positive field price_usd is not positive"},
    {"code": "unseen_category",
     "detail": "field currency has unseen categorical value"}
  ],
  "admissible_recovery_tools": [
    "get_stock_yahoo_finance",
    "get_stock_alpha_vantage"
  ]
}
\end{jsonbox}

The frozen confirmation uses the compact schema v1 above. The ToolMaze control
studies use receipt schema v2 (ledger field
\texttt{receipt\_schema\_version} value
\texttt{tool\_property\_path\_\allowbreak
expected\_observed\_first\_v2}), which
carries the paper's five-tuple witness. A real v2 receipt:

\begin{jsonbox}{Receipt, schema v2 (control studies)}
{
  "status": "inconsistent",
  "violation": {
    "tool": "get_weatherapi_alert_card_native",
    "property": "learned_type_mismatch",
    "path": "celsius_now",
    "expected": ["integer"],
    "observed": "string"
  },
  "admissible_recovery_tools": [
    "get_weatherapi_alert_card_native",
    "get_visualcrossing_forecast_chip_native"
  ]
}
\end{jsonbox}

\noindent\begin{minipage}{\columnwidth}
\paragraph{Generic-caution text (specificity control).}

The generic arm shares the detector, envelope, timing, and recovery
affordances, and replaces only the violation witness with one frozen
string:

\begin{jsonbox}{Generic-caution receipt (specificity control)}
{
  "status": "inconsistent",
  "violation": {
    "property": "generic_caution",
    "message": "This result may be unreliable; verify it before relying on it."
  },
  "admissible_recovery_tools": ["..."]
}
\end{jsonbox}

The decomposition controls
(Section~\ref{sec:decomposition-supp}) reuse this envelope. The warned
arms of the matched-trigger placebo and the full always-warn arm are
byte-identical to it; the stripped arm removes only the affordance key:

\begin{jsonbox}{Stripped always-warn envelope (decomposition study)}
{
  "status": "inconsistent",
  "violation": {
    "property": "generic_caution",
    "message": "This result may be unreliable; verify it before relying on it."
  }
}
\end{jsonbox}
\end{minipage}

\noindent\begin{minipage}{\columnwidth}
\paragraph{AppWorld and $\tau$-bench advisory text.}

Both ports append a plain-text block to the observed result. The
$\tau$-bench template (AppWorld differs only in naming its source ``its
public response contract'' and referring to ``documented APIs''):

\begin{promptbox}[breakable=false,top=2pt,bottom=2pt,
before upper={\footnotesize\ttfamily\raggedright}]{$\tau$-bench advisory template (appended to the raw result)}
[OUTCOME CONTRACT ADVISORY]\\
The observed result from `<function>` is inconsistent with a known
invariant of the retail database.\\
Violation: <code> at `<path>`; expected <expected>, observed
<observed>.\\
This advisory is nonbinding. Treat the result as unreliable and verify
or recover using the available documented tools and task context before
relying on it. All actions remain available.
\end{promptbox}
\end{minipage}

\paragraph{Fired invariant classes.}

Eight violation classes fire across the frozen ToolMaze ledgers, each
shown with a real fired instance:

\begin{itemize}\small
\item \texttt{nonpositive\_value}: nominally positive field
  \texttt{price\_usd} is not positive.
\item \texttt{unseen\_category}: field \texttt{currency} has unseen
  categorical value.
\item \texttt{learned\_type\_mismatch}: field \texttt{rate} has
  unexpected type.
\item \texttt{learned\_echo\_mismatch}: field \texttt{flight\_number}
  differs from call argument.
\item \texttt{missing\_learned\_field}: missing nominal field
  \texttt{display\_currency}.
\item \texttt{unexpected\_field}: field absent from nominal traces:
  \texttt{error}.
\item \texttt{explicit\_error}: error-bearing field(s): \texttt{error}.
\item \texttt{affine\_relation\_broken}: fields \texttt{cart\_subtotal}
  and \texttt{item\_count} violate a nominal relation.
\end{itemize}

\section{Organic Tool Traffic}
\label{sec:organic-supp}

Every paired-completion study in the main paper injects faults, because that
comparison requires a known fault. This section drops the agent and the counterfactual
and asks a narrower question that organic data can answer: how often does the
unmodified detector fire on real recorded tool responses?

\paragraph{Corpus and protocol.}
We use the response cache released with StableToolBench (Guo et al., 2024;
arXiv:2403.07714), which records real RapidAPI responses collected for
ToolBench (Qin et al., 2024); both are Apache-2.0. We keep responses the cache marks successful whose payload
is a JSON object, and we keep tools with at least 45 such responses. That leaves
427 tools and 53,078 responses. Per tool we mine contracts from 45 responses and
hold out 20, repeating over five seeds. The detector, the miner, and the
invariant classes are unchanged from the main paper's method section. Nothing is
injected and no fault is authored.

\begin{supptable}
\centering
\small
\begin{tabular}{lrr}
\toprule
Violation class & Firings & Share \\
\midrule
\texttt{unexpected\_field}      & 364 & 61.7\% \\
\texttt{missing\_learned\_field} &  78 & 13.2\% \\
\texttt{explicit\_error}        &  69 & 11.7\% \\
\texttt{learned\_type\_mismatch} &  53 &  9.0\% \\
\texttt{unseen\_category}       &  12 &  2.0\% \\
\texttt{learned\_echo\_mismatch} &   6 &  1.0\% \\
\texttt{nonpositive\_value}     &   6 &  1.0\% \\
\texttt{affine\_relation\_broken} &   2 &  0.3\% \\
\bottomrule
\end{tabular}
\caption{The detector fires on 0.80\% of organic responses, and firings
concentrate in 54 of 427 tools. Violation classes summed over five seeds on
held-out real RapidAPI responses; 6,860 responses scored per seed, 41 to 65
flagged. Field-shape classes dominate; the value and relation classes that
carry the main paper's effects fire rarely on this corpus.}
\label{tab:organic-supp}
\end{supptable}

\paragraph{Result.}
The detector flags 0.80\% of held-out responses (41 to 65 of 6,860 per seed,
range 0.60\% to 0.95\%). For comparison, the clean-P0 control in the main paper
puts the false-positive rate at 1.39\% on synthetic clean workflows, so the
synthetic estimate is not optimistic relative to this corpus. Firings are
concentrated rather than diffuse: 54 of 427 tools account for all of them.

\paragraph{What the firings look like.}
We do not adjudicate every flag, and field-shape deviations on a public API are
often benign. Two recurring patterns are worth showing. The first is an error
delivered inside a response the transport layer reports as successful, which is
the failure mode the paper is about:

\begin{quote}\small\ttfamily
airportflights $\rightarrow$ \{"Error": "invalid ident, please provide a valid
ident. You can use the airport search to determine the right one."\}
\end{quote}

The second is an expected payload that is simply absent, as when
\texttt{coins\_get\_markets} returns \texttt{\{\}} where the mined contract
expects a \texttt{data} field.

\paragraph{Scope.}
This is a detection measurement, not an effectiveness one. There is no agent,
no receipt, and no counterfactual, so nothing here speaks to recovery or
completion in deployment. The responses are cached and filtered by
StableToolBench rather than collected live. What the section establishes is narrower. Contracts mine cleanly from
organic traffic. The detector's firing rate there is consistent with the
synthetic clean-P0 estimate. Silent failures of the kind the paper targets
are present in real tool traffic, without anyone injecting them.

\section{Construction Cost Detail}

ToolMaze: 400 unique nominal workflows produce 1,800 workflow-fold inputs
across five cross-fitted registries. The build materializes 15,551 candidate
invariants and accepts 15,041 (96.7\%). Invariant classes include per-field
type constraints, echo fields ($a[f]=y[f]$), positivity ($y[f]>0$),
categorical domains, affine relations, and temporal orderings.

AppWorld: 11 released API-documentation files yield 473 candidate read
contracts, of which 149 (31.5\%) are accepted after recursive schema
validation.

$\tau$-bench: the public \texttt{orders.json} database yields categorical
(status, transaction\_type), range (item price $> 0$), and affine
(net payment conservation) contracts.

All builds are deterministic with zero model or API calls. Historical human
authoring and inspection time was not contemporaneously recorded and is
reported as unknown.

\section{Reproducibility Details}

\paragraph{Models and access.}
DeepSeek V4 Flash and V4 Pro via the direct DeepSeek API. Qwen 3.7 Plus and
3.7 Max via OpenRouter. The separate MiniMax M3 replication uses OpenRouter
pinned to the MiniMax-hosted FP8 route with fallbacks disabled. All ToolMaze
runs use temperature zero and extended reasoning disabled.

\paragraph{Token usage and cost.}
The ToolMaze confirmation uses 24.81 million input tokens and 0.906 million
output tokens; its all-cache-miss list-price upper bound is \$15.46.
The AppWorld development study uses 0.539/0.741 million cache-miss input
tokens, 7.56/7.40 million cache-hit input tokens, and 0.332/0.328 million
output tokens for Flash/Pro, at \$1.204 total list price. Its held-out
\texttt{test\_normal} study uses 0.724 million cache-miss input tokens,
9.754 million cache-hit input tokens, and 0.383 million output tokens, at
\$0.345 total list price. The method comparison costs \$14.926. The
always-verify and clean-P0 controls cost \$2.730 and \$2.372, respectively.
The MiniMax M3 replication costs \$2.18. The frozen $\tau$-bench factorial
costs \$0.43. The matched-trigger placebo study costs \$5.73 and its
stripped-envelope companion \$4.36.

\paragraph{Artifact mapping.}
The anonymous artifact's \texttt{MANIFEST.md} and
\texttt{ARTIFACT\_INDEX.md} map each paper result to its sanitized manifest,
aggregate report, and analysis code. Row-level provider responses and protected
benchmark observations are intentionally excluded.

\noindent\textbf{Selected study status and analysis units.}\par

\begin{supptable}
\centering
\scriptsize
\setlength{\tabcolsep}{2pt}
\begin{tabular}{@{}p{.25\columnwidth}p{.20\columnwidth}p{.46\columnwidth}@{}}
\toprule
Study & Status & Unit and aggregate \\
\midrule
ToolMaze primary & frozen & 80 shared tasks; four-model aggregate \\
MiniMax replication & frozen separately & 80 paired tasks; per-run analysis \\
Reasoning ablation & separate hashes & 80 shared tasks; paper aggregate \\
Specificity / clean & separate hashes & 57 shared workflows; per-run joint analyses \\
Matched placebo & frozen & 57 fresh workflows; two-tier joint \\
Stripped always-warn & frozen & same 57 workflows; three contemporaneous arms \\
AppWorld dev & development & 19 families; joint task analysis \\
AppWorld held-out & frozen, Flash run & 32 families; Flash analysis \\
$\tau$-bench factorial & frozen & 50 tasks; two fault contrasts per task \\
\bottomrule
\end{tabular}
\caption{Selected study status and analysis units. The artifact's
\texttt{STUDY\_INVENTORY.csv} lists all 22 studies and 4,683 episodes.
Separately hashed studies are never
pooled. The held-out AppWorld manifest planned two tiers, but only Flash was
executed. MiniMax is never pooled into the primary ToolMaze aggregate; the
$\tau$-bench secondary aggregate clusters both contrasts by task.}
\label{tab:study-status-supp}
\end{supptable}

\noindent\mbox{\textbf{Aggregate regeneration.}}\par\nobreak
\texttt{paper\_aggregate\_analysis.py} validates all four frozen ToolMaze
ledgers in the research workspace, reconstructs the shared-task effect and
bootstrap interval, and independently reproduces the 484-event recovery-action
counts. The review artifact packages the resulting aggregate, analysis code,
and independent headline-value and cross-study validators, but not row-level
provider responses. The aggregate explicitly excludes the retired
substitute-availability post-hoc because its classification requires the
unavailable external task catalog.

For AppWorld, protected API-observation ledgers remain ignored; the packaged
Flash aggregate and report contain only family-level counts, injection rates,
usage, and hashes. The $\tau$-bench analysis validates the frozen
$50\times5$ factorial grid and records task-clustered inference separately
from the prespecified fault-stratified tests. Finally, \texttt{submission.tex} is a
flattened, bibliography-inlined source generated from the modular manuscript;
the build checks that it has the same page count as \texttt{main.pdf}.